\documentclass[letterpaper, 10 pt, conference]{ieeeconf}  

\IEEEoverridecommandlockouts                              

\usepackage{subcaption}
\usepackage{graphicx}
\usepackage{amsmath}
\usepackage{amsfonts}
\usepackage{amssymb}
\usepackage{mathtools}
\usepackage{bm}
\usepackage{wrapfig}
\usepackage{gensymb}

\usepackage{fixme}
\fxsetup{status=draft, theme=color}
\definecolor{fxtarget}{rgb}{0.8000,0.0000,0.0000}
\definecolor{fxnote}{rgb}{0.0000,0.0000,0.8000}

\definecolor{MyBlue}{rgb}{0,0.2,0.8}
\let\labelindent\relax
\usepackage{enumitem}
\title{\LARGE \bf
Contextual Reinforcement Learning of\\ Visuo-tactile Multi-fingered Grasping Policies
}

\author{Visak Kumar$^{*1}$, Tucker Hermans$^{*2}$, Dieter Fox$^{3}$, Stan Birchfield$^{3}$ and  Jonathan Tremblay$^{3}$\\
\thanks{$^{*}$Work performed during at NVIDIA}%
$^{1}$School of Interactive Computing at Georgia Institute
of Technology, {\tt\small visak3@gatech.edu}\\%
$^{2}$ Robotics Center and the School of Computing at University of Utah,
        {\tt\small thermans@cs.utah.edu}\\%
$^{3}$ NVIDIA,
        \texttt{\small \{dieterf, sbirchfield, jtremblay\}@nvidia.com}%
}

\begin{document}

\maketitle
\thispagestyle{empty}
\pagestyle{empty}

\begin{abstract}

Using simulation to train robot manipulation policies holds the promise of an almost unlimited amount of training data,
generated safely out of harm's way.
One of the key challenges of using simulation, to date, has been to bridge the reality gap,
so that policies trained in simulation can be deployed in the real world.
We explore the reality gap in the context of learning a contextual policy for multi-fingered robotic grasping.
We propose a Grasping Objects Approach for Tactile (GOAT) robotic hands, learning to
overcome the reality gap problem.
In our approach we use  human hand motion demonstration to initialize and reduce the
search space for learning.
We contextualize our policy with the bounding cuboid dimensions of the object of interest, which
allows the policy to work on a more flexible representation than directly using an image or point cloud.
Leveraging fingertip touch sensors in the hand allows the policy to
overcome the reduction in geometric information introduced by the coarse bounding box, as well as pose estimation uncertainty.
We show our learned policy successfully runs on a real robot without
any fine tuning, thus bridging the reality gap.

\end{abstract}

\section{INTRODUCTION}\label{sec:intro}

Enabling robots to autonomously grasp object of varying shape and size with multi-fingered hands stands as a fundamental challenge necessary to produce more general manipulation skills such as pick-and-place tasks, human handover, and dexterous tool use.
Classical solutions to this problem take a model-based planning and control approach. A typical pipeline estimates the object pose, given either a 3D point cloud or mesh of the object, then plans a set of contact locations and hand configuration to define the grasp, and finally generates a motion plan to reach and grasp the object.
Such systems are sensitive to perception and calibration errors and often require significant computational time to plan and execute~\cite{Dai2015}.
Such issues might cause the system to misbehave and fail to grasp the object.

In this work we propose to overcome these constraints by learning a policy to grasp objects of varying geometry and scale with a multi-finger gripper using deep reinforcement learning (RL).
A few important challenges arise in formulating the multi-fingered grasping problem as an RL problem.
First, how to cope with the relatively high dimension of the multi-fingered hand's configuration space in order to effectively explore the space of possible grasping policies? Second, how should the learner represent the object to be grasped in a way that can effectively generalize across objects of varying shape, while still being succinct enough to train efficiently? Third, how can we learn such a policy purely in simulation with no need to fine tune the policy for use in the physical world?

In order to efficiently search over the high-dimensional space of grasping policies, we leverage recent advancements in camera-based human hand pose estimation~\cite{iqbal2018hand} and imitation learning~\cite{peng2018deepmimic} to provide human grasping demonstrations from an RGB camera. We use these grasping demonstrations as a component in our reward function, providing a prior for preferred grasping trajectories to the learner in simulation.

We address the problems of object representation and sim-to-real transfer by proposing a bounding-box based object representation. We extract the location of the 8 vertices of the cuboid enveloping the object to provide the object's pose, general shape, and size as a context variable to the policy. Using these keypoints explicitly as a context variable and training over a variable set of object shapes enables our policy to adapt to different block-shaped objects upon deployment without the need for further training.

However, this does not enable the robot to robustly compensate for object geometries, such as cylinders or cones, not tightly captured by bounding box. As such we additionally make use of tactile sensing to provide contact information as part of the robot's state. This enables the policy to learn that making--and maintaining--contact is necessary for grasping. This has the further benefit of aiding in bridging the sim-to-real gap, where tactile sensors on the physical robot compensate not only for object shape mismatch but also localization and calibration error from visual sensing. We deploy our final learned policy onto a real world system where visual input to the policy
comes from an RGB pose estimator~\cite{tremblay2018corl:dope}
and the contact information is retrieved from BioTac tactile sensors.


Our approach differs from many recent RL for sim-to-real tasks which attempt to overcome poor parameterization of the system dynamics or object and environment appearance by learning policies robust to high variation in visual sensing~\cite{Tobin2017DomainRF,Tobin2018DomainRA,Fereshteh}. We take an alternative approach of abstracting away the uncertain object appearance and geometry into a succinct set of geometric features. To account for the coarse approximation these features induce, we leverage tactile sensors in the robot's fingertips to observe contacts explicitly as part of our state. This differs also from standard approaches to grasp learning where richer visual features are leveraged to understand the object geometry at a relatively high resolution; where these features are either learned~\cite{varley2015generating,lu2017planning,lu2019modeling} or hand-crafted~\cite{osa-ar2018-grasping,kopicki2016one}.

This work makes the following 
contributions:
\begin{itemize}[leftmargin=10pt,labelindent=10pt,topsep=0pt]
\item We present a system that leverages human demonstrations of grasping, reinforcement learning and
  sim-to-real to accomplish a multi-finger grasp task on a real-world system.
  We demonstrate that our system generalizes to unseen shapes in the real-world without any fine tuning.
\item We introduce a novel approach to fusing visual and tactile information in learned grasp policies, using 3D keypoints for context variables encoding object shape and binary contact signals within our object state.
  This allows our policy to reason about the object size and orientation implicitly
  creating a versatile policy that can adapt locally by leveraging the sensed contact information.
\end{itemize}

\vspace{1ex}
We provide empirical results demonstrating that benefits of our various contributions. 
We show that our keypoint representation coupled with tactile feedback can successfully grasp objects of varying shape not seen in training. We additionally quantify the benefits of using human hand grasping demonstration motions in learning a multi-fingered grasping policy. We show that our learned policy achieves comparable results to a hand-engineered policy on a real-word, physical robot without any fine tuning.
We further demonstrate the ability to grasp with varying grasp styles simply by changing the human demonstrations provided during training. We will release our dataset of captured human hand motions used to teach our robot to grasp with style upon publication.


 \section{Method}\label{sec:method}
We now present the details of our approach to learning grasping policies for multi-fingered hands. We begin with a brief background of contextual policy search for reinforcement learning. We then give the specifics of how we encode the grasping problem into this contextual policy search framework. Following this we discuss how we learn policies informed from demonstration using RL. We conclude the section by describing how the policy is deployed on the physical robot.

\subsection{Background: Contextual Policy Search}
We formulate the task of multi-finger grasping as a contextual policy search problem~\cite{Kober2011}. This differs from the classic Markov Decision Process (MDP)~\cite{Sutton1998a} in that the agent (robot) observes some context variable \(\bm{\kappa}\) at the beginning of the episode which parameterizes the reward function $r: \mathcal{S} \times \mathcal{A} \mapsto \mathbb{R}$; where $\mathcal{S}$ and $\mathcal{A}$ define the state action spaces respectively. The objective of the contextual policy search problem remains the same as standard reinforcement learning, namely to find a policy $\pi: \mathcal{S} \mapsto \mathcal{A}$, that maximizes the expected accumulated reward, conditioned on the observed context \(\bm{\kappa}\):
\begin{equation}
    J(\pi_{\bm{\theta}}) = \mathbb{E}_{\mathbf{s}_0, \mathbf{a}_0, \dots, \mathbf{s}_T} \sum_{t=0}^{T} \gamma^t r(\mathbf{s}_t, \mathbf{a}_t; \bm{\kappa}),\nonumber
\end{equation}
 where $\mathbf{s}_0 \sim p_0$, $\mathbf{a}_t \sim \pi_{\bm{\theta}}(\mathbf{s}_t; \bm{\kappa})$, and $\mathbf{s}_{t+1}=\mathcal{T}(\mathbf{s}_t, \mathbf{a}_t)$.
The remaining components of the MDP also exist in our problem formulation, specifically $\mathcal{T}: \mathcal{S} \times \mathcal{A} \mapsto \mathcal{S}$ is the transition function, $p_0$ is the initial state distribution and $\gamma$ is the discount factor. We additionally make explicit the policy parameters \(\bm{\theta}\) which we seek to learn through roll-outs of the system.

\begin{figure}
\centering
\includegraphics[width=0.65\linewidth]{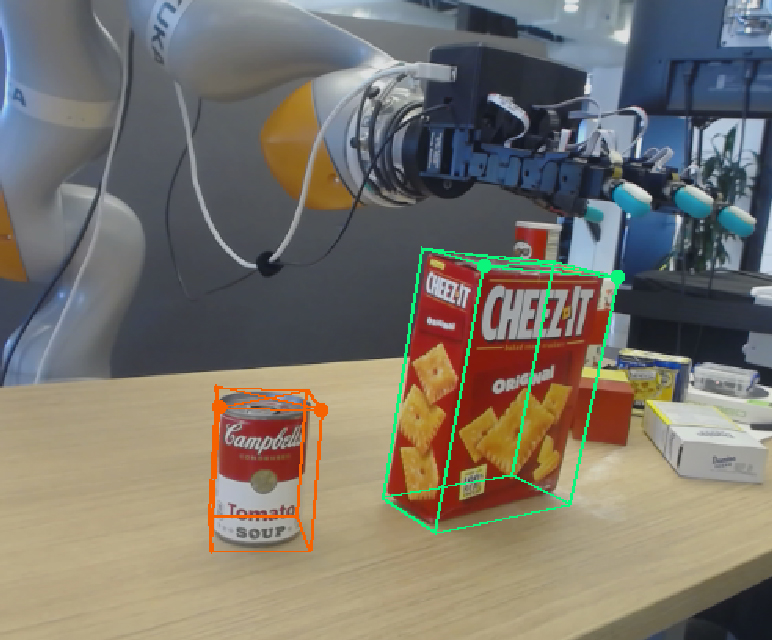}
\caption{\label{fig:keypoints}Visualization of the context variables keypoints for two objects. Note the pose estimation error for both objects and the mismatch in object shape for the soup can.}
\end{figure}

\subsection{Grasping as Contextual Policy Search}
We define the context variables, \(\bm{\kappa}\), for our multi-fingered grasping problem as the keypoints of a bounding box surrounding the object of interest at its pose at the beginning of the episode (see Fig.~\ref{fig:keypoints}).
This defines a low dimensional feature representation to encode the object geometry,
there are several ways to infer these features at runtime such as using pose estimation of known objects~\cite{tremblay2018corl:dope}.
By providing this information of the object's pose only at the beginning of the trial, we remove the need to explicitly track the object during the execution.
We believe this to be an advantage as stably tracking the object, even when a known model exists, remains challenging, because of the inevitable (partial) occlusion of the object caused by the hand interacting with it.
Since the initial estimate may be inaccurate and the object will likely move during execution, we provide binary contact information for each robot fingertip as part of the robot's state space.

In simulation we can directly observe contacts using the model of the robot and object.
On the physical system we estimate contact using the pressure sensors of the BioTac sensors embedded in each fingertip.
In addition to localizing the object, we hypothesize that contact information provides an extremely useful signal in learning stable grasps that can generalize across different objects geometries.
The state space includes the Cartesian palm location denoted by $P_{xyz} \in \mathbb{R}^{3}$
and orientation $u \in \mathbb{R}^{4}$ all defined in the robot base frame,
joint positions and velocities of the 16 DOF four-fingered hand represented as $q_h$ ($\mathbb{R}^{16}$) and $\dot{q_h}$
($\mathbb{R}^{16}$) and contact vector $c$ which contains binary contact information about the four fingertips
($\mathbb{Z}_2^{4}$).
This results in final state space of dimension \(43\). 
The context variable $\kappa$ is \(24\) dimensional, it contains the Cartesian $x,y,z$ ($\mathbb{R}^{3}$) locations of each corner of a cuboid in the robot base frame.
We define the robot action space as the desired Cartesian hand pose and the desired joint positions of the fingers. As such our action space has 22 dimensions.


\subsection{Reward Function}

The task of reaching and grasping a wide range of objects with a multi-fingered hand is not trivial and
as such we introduce reward terms to overcome several different challenges. We present each reward term in turn below; we define the final reward as the sum of these terms with weights selected such that each component has relatively equal scale.

\textbf{Hand location with respect to the object.} The first reward component encourages moving the palm of the hand close enough to the object to enable contact.
Assuming a valid object pose estimate, keypoint locations of the object $\kappa$ are computed
in the robot base frame.
We use the average of the 4 keypoint locations on the top surface of the object, denoted $\kappa_{\textit{offset}}$,
to compute the following reward:
\begin{equation}
\mathcal{R}_{\textit{pos}} = \exp{\left\{-w_1||P_{xyz} - \kappa_{\textit{offset}}||\right\}}\label{eq:r_pos}
\end{equation}
\textbf{Hand motion.} The second reward component serves to focus the policy search on likely to work motions in order to overcome the relatively high-dimensional configuration space of multi-fingered hands (16 DOF for our Allegro hand).
To tackle this issue, we use human demonstrations, captured from a hand pose estimator~\cite{iqbal2018hand}, as useful prior information for policy learning.
This, however, introduces another concern as the kinematic structure of the human hand is different from the robot's. 
Since we know the values of the kinematic link lengths of the Allegro hand and the human hand from which demonstrations are generated, we perform a simple re-scaling of the data to fit the robot hand dimensions. In addition, we only reward the policy when the robot's fingertip locations $q^{e}$ track the fingertip locations obtained from the human hand pose estimator $\hat{q}^{e}_{i}$. The purpose of the demonstrations is not to provide an accurate trajectory for the fingers to follow, but to reduce the search space of the policy.
\begin{equation}
\mathcal{R}_{\textit{hand}} = \exp{\left\{-w_2\sum_{i=1}^{4}||q^{e}_{i} - \hat{q}^{e}_{i}||\right\}}\label{eq:r_hand}
\end{equation}
\textbf{Task success:} Once the robot grasps the object, we reward the policy if it can successfully lift the object to a position, $p^o_y$, above its starting location, $p^s_y$:
\begin{equation}
    \mathcal{R}_{\textit{lift}} =
    \begin{cases}
      w_3, & \text{if}\ p^o_y > p^s_y \\
      0, & \text{otherwise}
    \end{cases}\label{eq:r_lift}
\end{equation}
\textbf{Contact.} Our reward function also encourages the robot to make fingertip contact with the object. We hypothesize that contact information greatly improves the ability to learn a stable grasping policy across objects of varying size and geometry. Here we define variable \(c_i\) to have value 1 if fingertip \(i\) is in contact and 0 otherwise:
\begin{equation}
  \mathcal{R}_{\textit{contact}} = \sum_{i=1}^4 c_i
\label{eq:r_contact}
\end{equation}
The goal of our control policy is to generalize to objects of different geometry. 
The structure of our reward function with multiple terms reflect this goal, {\em e.g.,} touch sensing and cuboid keypoints. 
In our experiments, we found that a binary/sparse reward for a task involving a multi-fingered robot to reach and grasp an object is not feasible, the reward is too sparse to learn anything. 
We assume in our experimental set up that the hand starting location is near the object of interest. 

\subsection{Training Details}
We use the proximal policy optimization (PPO) \cite{Schulman2017ProximalPO} algorithm to learn the policy.
We represent the policy as a multi-layered perceptron (MLP) with 2 hidden layers containing 128 neurons each.
During training, at the beginning of each rollout we generate a new cuboid object with dimensions uniformly sampled from a pre-specified range, we estimate the keypoints of the object---sampled noise is added to the keypoint locations to simulate sensor noise present in the physical system---and pass them as context to the policy.
The keypoint values then remain the same throughout that rollout.
Since we wish to deploy the policy learned in simulation on a real robot,
we apply domain randomization on the objects to account for the discrepancy between the simulator and physical world. In addition to keypoint location noise, we add uniform noise to the object mass, friction coefficients between the fingers and object, PD gains of the robot, and damping coefficients of the robot joints.
The range of the uniform distribution was manually specified based on initial results on the robot. Our method takes about $6$ hrs of training time with four threads on an i7 collecting $1.2e6$ samples across $500$ iterations. These numbers are consistent across four different seeds. 

\subsection{Keypoint Parameter Adaptation for Novel Geometries}\label{sec:keypoint-adaptation}
A primary goal of our approach is to learn a policy that generalizes to objects of non-cuboid shapes not seen during training.
In essence, a new object implies a new context for the policy.
While we can use the bounding box of a novel object to extract the keypoints defining the context variables, we find that this does not work well for objects with shape that significantly differs from the bounding box. As such, we propose optimizing over the context variables in order to find values which will enable the pre-trained policy to succeed. Importantly, we remove the restriction that the keypoints define a recta-linear box allowing them to take any point in 3D.

Given a policy trained in simulation over a uniform distribution of contexts, when presented with a new object we fix the policy network and search over the context variables using CMA-ES. We initialize the keypoints using the object bounding box. We evaluate the objective function by running a rollout in simulation and provide the height reached by the object once lifted as a continuous reward for the planner to maximize. 
In each iteration, there are about 5 rollouts of the policy, which means that about 65-70 trajectories on the new object to fine tune the policy.
This whole process takes about 20 min of compute time. 
We examine the benefit of this adaptation in Section~\ref{sec:parameter-adaptation-results}.

\section{Experiments}\label{sec:experiments}
We evaluate our method both in simulation and on the real robot.
In these experiments we answer the following overarching questions.
First, how important is hand demonstration data to learn an effective policy?
Second, how does including contact information change the effectiveness of the grasp?
Third, how sensitive is the policy learning to the object feature representation?
And fourth, can our policy successfully transfer to a real robot without adaptation?

As such this section is organized as follows: We first discuss the implications of our state representation and reward functions by comparing GOAT to different
baselines.
Then we quantify how parametrization search over our keypoint representation can improve the learned policy's performance.
In addition to these experiments, we also show that using our method we can grasp objects with 6 different styles and evaluate the effectiveness of the different grasp styles.
We conclude this section by showing real-world experiments on the robot.


\subsection{Comparison Methods}
In order to evaluate the proposed method, we compare it to three baselines:

\textbf{Baseline 1.} The policy does not use any contact information; 
we hypothesize that local contact information is important in adapting to non-cuboid shapes and for identifying stable grasps once the robot hand makes contact with the object.

\textbf{Baseline 2.} We include contact information, however, we do not reward the policy for tracking the human hand demonstrations---{\em i.e.}, we set the weight in Eq.~(\ref{eq:r_hand}) to 0. 
We would like to test the importance of demonstration data in learning in this high-dimensional action space, which, combined with sparse nature of the reward, makes it a difficult reinforcement learning problem.

\textbf{Baseline 3.}  We change the context variable $\kappa$ to a single 6-DoF pose vector of the object's center. 
This tests our hypothesis that using keypoint information as the context variable provides a coarse representation of the object geometry enabling the policy to adapt to objects of varying shape.

To compare the effectiveness of our method to that of the policies trained using the baseline methods we perform two different tests.
First, we generate 100 random objects unseen by the policies during training and test grasps for each object from 5 random poses on the table. We compare the number of successful grasps out of these 500 resulting trials.

\begin{figure}
\centering
\includegraphics[width=0.7\linewidth]{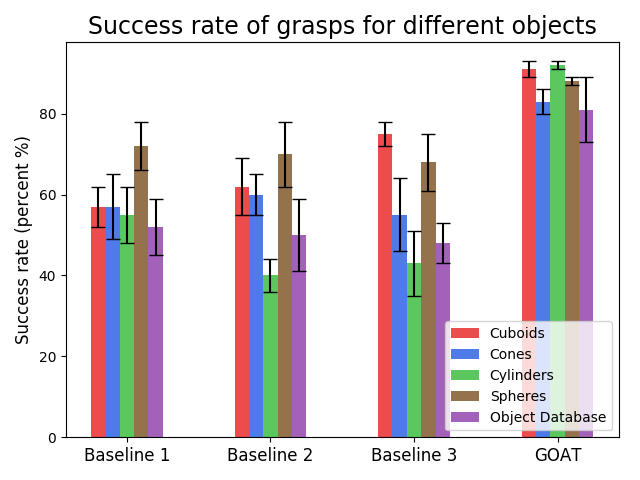}
\caption{Grasp success rate of trained policies in simulation. The experiment was done for four different seeds.} \label{fig:Basea}

\end{figure}

\begin{figure}
\centering
\includegraphics[width=0.7\linewidth]{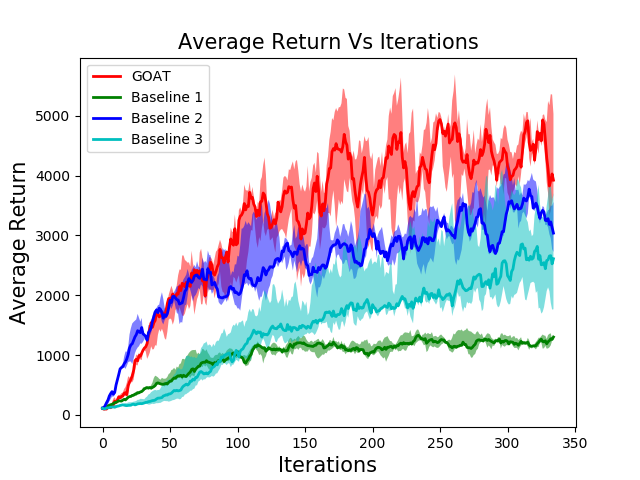}
\caption{
Average reward achieved during learning for the different methods averaged over four initial seeds for the cuboid category.
} \label{fig:Basec}
\end{figure}

Figure~\ref{fig:Basea} illustrates the number of successful grasps achieved by each method on different object types.
Each bar represent the average of four different trained seed on a specific category. 
Object Database refers to the open source dataset of 3d objects grasp database~\cite{kappler2015leveraging} where we randomly used 20 objects.
We can clearly see that our proposed method outperforms all the baselines for the different object types. 
Interestingly the baselines all perform somewhat similarly and thus suggesting that our method provides the most detailed information 
for accomplishing this task. 
We also show the learning curves for average reward achieved by each method during training in Figure~\ref{fig:Basec} for the cuboid category. Learning curve results represent the average and variance over four different seeds. 
It is worth noting that the weighting of the reward function remains the same across all experiments.

\begin{figure}
\centering
\includegraphics[width=0.7\linewidth]{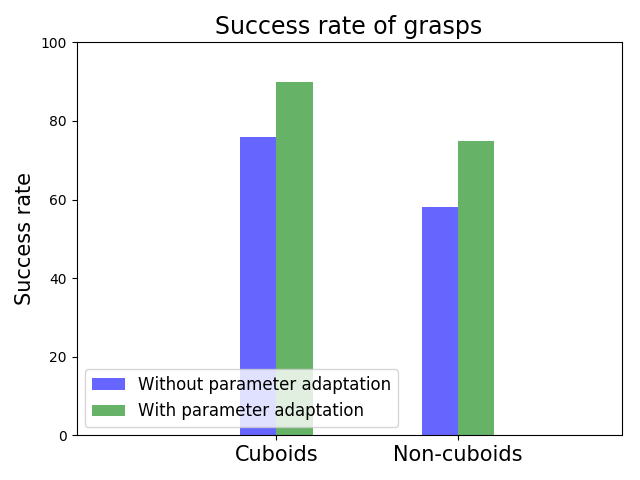}
\caption{Parameter adaptation on the keypoints to improve the performance of the policy. 
For both cuboid and non-cuboid shapes, parameter adaptation improves policy performance.} \label{fig:Adapa}
\end{figure}

\begin{figure}
\centering
\includegraphics[width=0.7\linewidth]{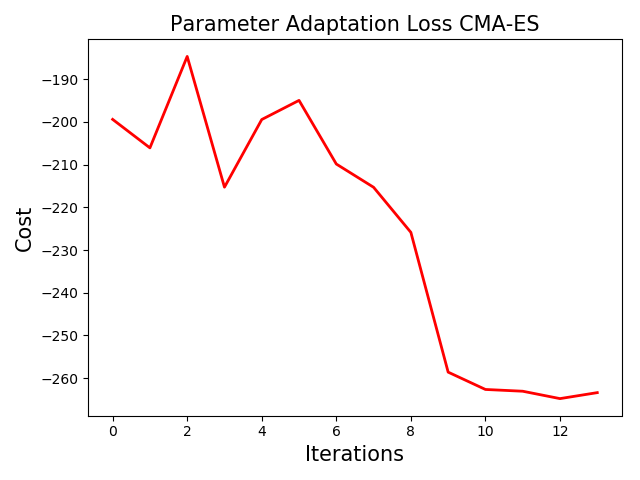}
\caption{CMA-ES optimization loss curve.  Convergence is achieved after 13 iterations.} \label{fig:Adapb}
\end{figure}

\begin{figure*}[h!bt]
\centering
\setlength{\tabcolsep}{0.2pt}
\renewcommand{\arraystretch}{0.7}
\begin{tabular}{c c c c c}
  \includegraphics[width=0.2\textwidth,height=3cm]{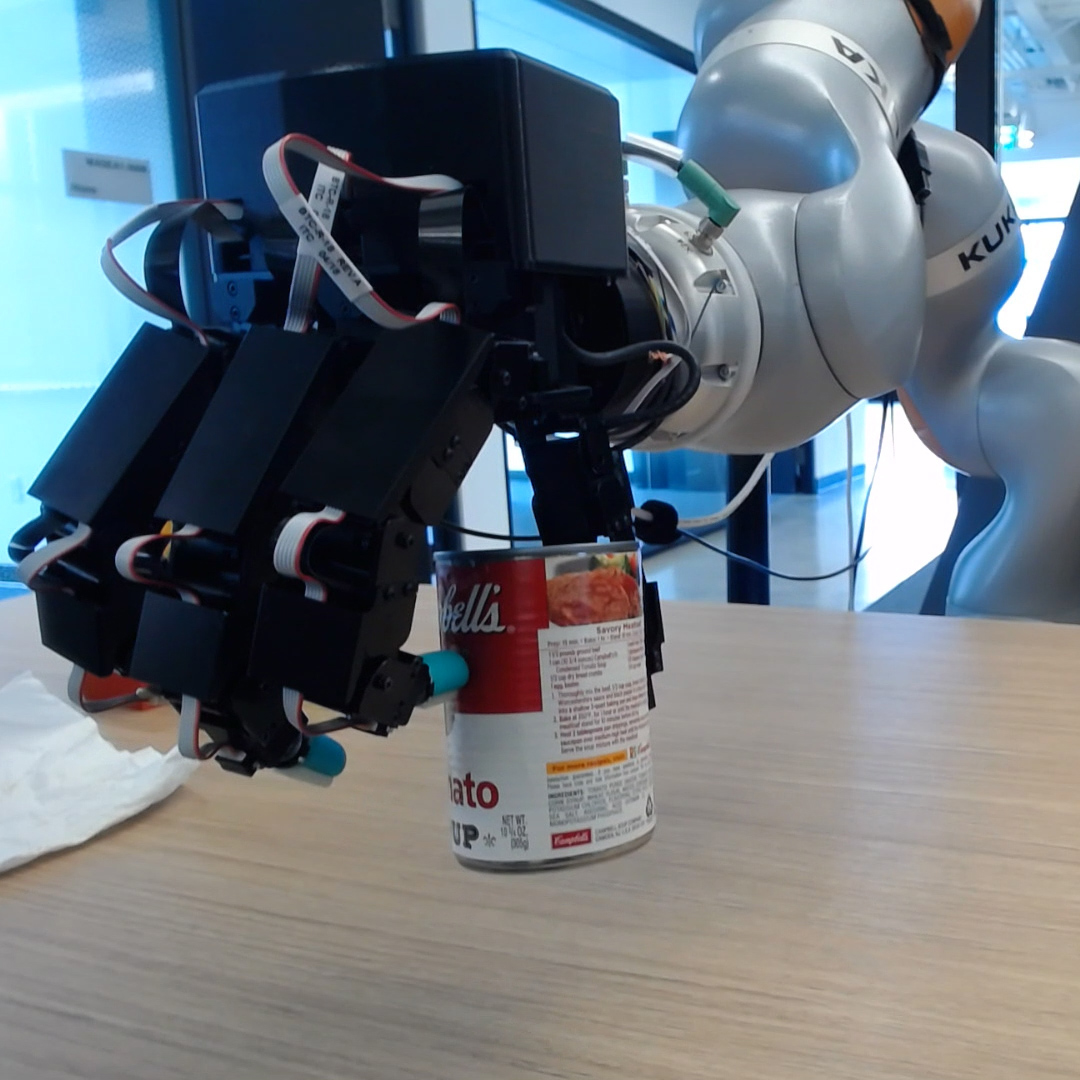}&
  \includegraphics[width=0.2\textwidth,height=3cm]{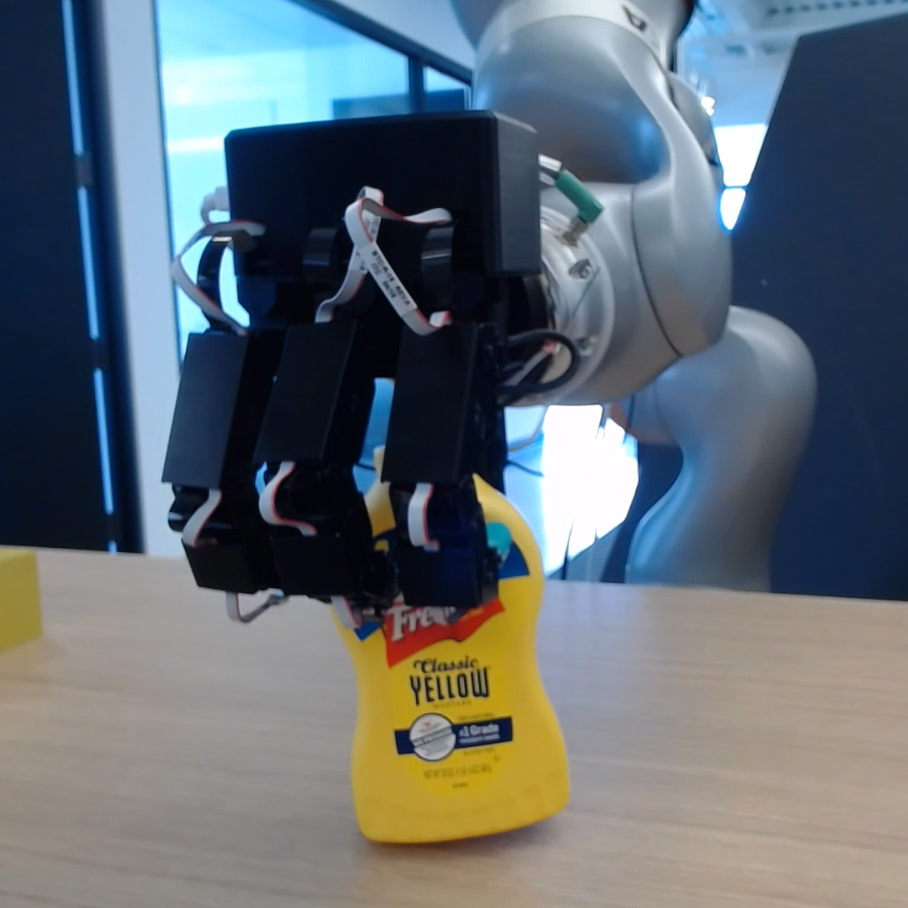}&
  \includegraphics[width=0.2\textwidth,height=3cm]{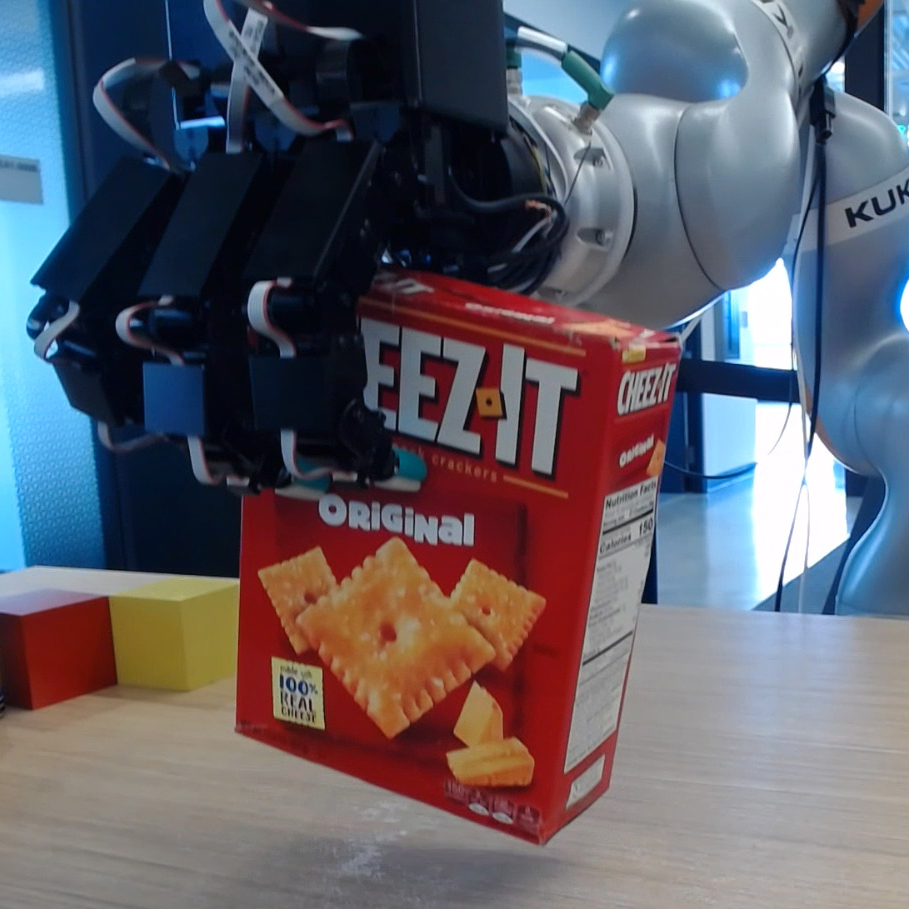}&
  \includegraphics[width=0.2\textwidth,height=3cm]{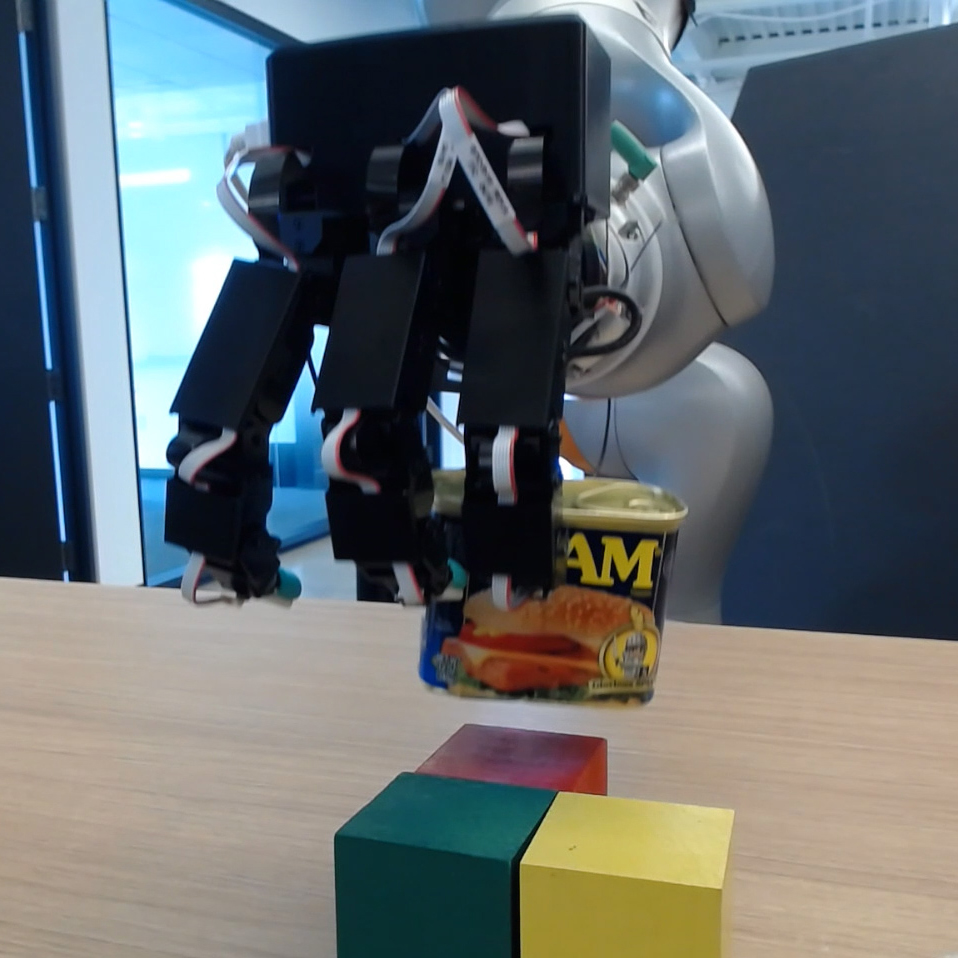}&
  \includegraphics[width=0.2\textwidth,height=3cm]{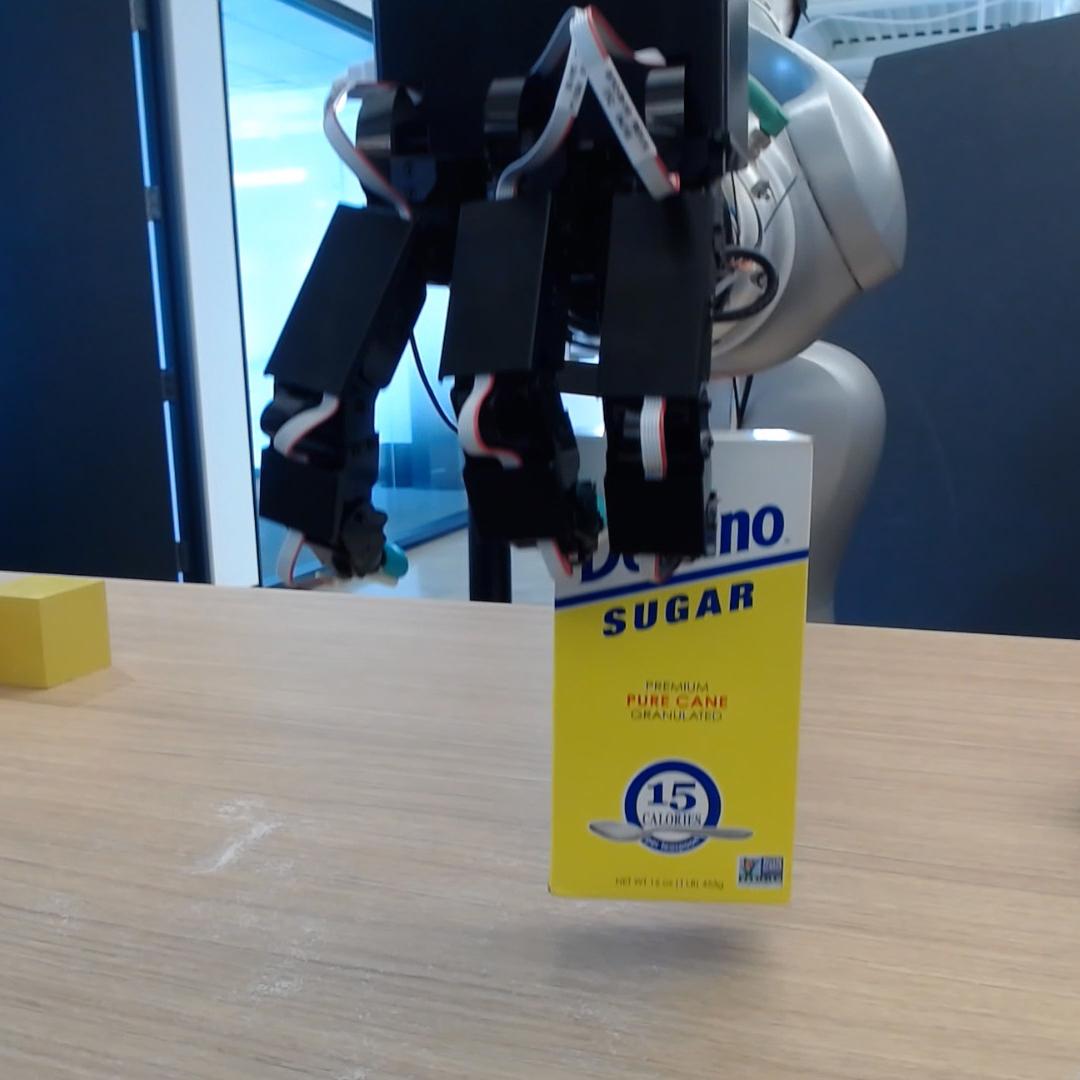} \\
\end{tabular}
\caption{Representative grasps generated by our policy executed on the physical robot.}
\label{fig:example-grasps-real}
\end{figure*}

\subsection{Parameter Adaptation experiments}\label{sec:parameter-adaptation-results}
In the previous experiment with unseen objects, we tested the trained policy with context parameters selected from the object bounding box provided by our simulator. We ran experiments to investigate the effect of keypoint adaptation approach presented in Section~\ref{sec:keypoint-adaptation}. Figure~\ref{fig:Adapa} shows the improvement in grasp success rate after parameter adaptation for both cuboid and non-cuboid objects. Figure~\ref{fig:Adapb} illustrates how the optimization loss reduces during the parameter adaption process. It takes on average \(\approx10\) iterations of the CMA-ES to identify keypoint inputs that enable the policy to pick up novel objects.

\begin{figure}
  \centering
  \includegraphics[width=\linewidth]{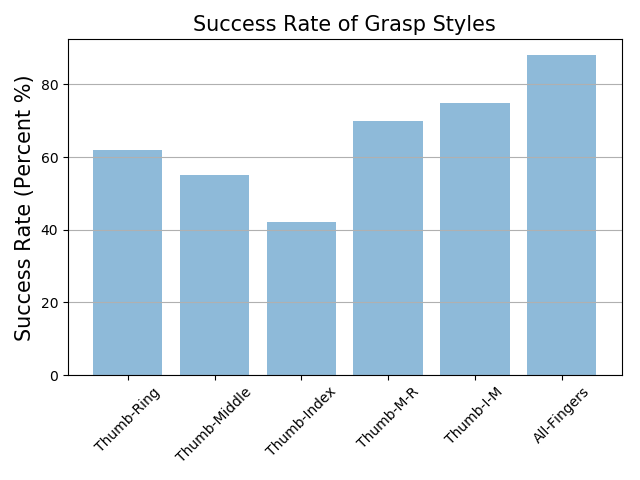}
  \caption{Success rates of different grasp styles. M, R and I refer to middle, ring and index finger respectively.}
  \label{fig:stylesr}
\end{figure}
\subsection{Grasping with Style}
To leverage the hand pose data made available by the hand pose estimator we learn different grasping styles. 
For the purpose of this experiment we define a grasping style as simple motions that the robot has to follow, 
{\em e.g.}, only using the thumb and the index finger for grasping.
Figure~\ref{fig:stylesr} illustrates the grasp success rate of each of the different styles. 
As expected, two fingered grasps are not as successful as with three or four fingered grasps. 
The objects used for this test were a mixture of 50\% cuboid and 50\% non-cuboid shapes.

\subsection{Real Robot}
The ultimate test for GOAT is whether the learned policy can be deployed onto a
real world robot.
We use an Allegro robotic hand with 4 BioTac sensors mounted on a 7-DoF Kuka LBR iiwa 7 R800 arm.
We use the pressure sensing on the BioTac to detect contact, and it is quite sensitive.  
We use DOPE~\cite{tremblay2018corl:dope} to localize the object and generate its bounding box keypoint locations.
%
%
We use the 5 objects DOPE can detect from the YCB dataset \cite{calli2015benchmarking}:
cracker box, meat, mustard, soup, and sugar box.
Other methods could be used here to fit a bounding box around the object, similar to \cite{lu2019grasp}, 
we could leverage point cloud sensing to fit a bounding box on points above the work surface assuming 
a non cluttered environment. 
During the experiment, the object was placed randomly within the robot's workplace five times
with a random in plane orientation
between $-30^{\circ}$ and $30^{\circ}$, where \(0^{\circ}\) means the object's axis is aligned with the robot base.
For each pose detection we sample a normal distribution with variance of 1 mm or 10 mm 
to perturb the object location. 
We consider a successful grasp if the object stays above the work surface for at least 5 seconds. 

We compared our method against a handwritten grasping policy, denoted baseline. 
Our baseline simply moves to a position 6 cm above the estimated center of the object. Once it reaches this location,
the hand begins closing its fingers towards the object. Each finger stops moving when it detects contact with the object. Once all fingers have touched the object the hand exerts more force on the object before lifting it up 7 cm.

\begin{table}[th]
\centering
\caption{Experiments showing GOAT performance on a real world against a hand tuned baseline.}
\label{tbl:realpick}
\resizebox{\columnwidth}{!}{%


\begin{tabular}{l|cc|cc}
                            &  \multicolumn{2}{c|}{noise = 0.001}& \multicolumn{2}{c}{noise = 0.01} \\
objects                     & baseline          & GOAT          & baseline          & GOAT         \\
\hline
cracker box                 & 5                 & 5             & 3                 & 5            \\
meat                        & 5                 & 5             & 2                 & 2            \\
mustard                     & 5                 & 3             & 3                 & 3            \\
soup                        & 3                 & 4             & 0                 & 1            \\
sugar box                   & 5                 & 5             & 4                 & 4            \\
\hline
all                         & 23/25             & 22/25         & 12/25             & 14/25        \\

\end{tabular}

}
\end{table}
Table~\ref{tbl:realpick} depicts our results, it shows that our method performs similarly to the
baseline under different noise levels.
The soup is quite a challenging object for performing a top grasp,
we were surprised to see our method moving its finger in such a way that it was looking for the object
and achieving stable grasp with the cylinder even though it was never trained on such physical object.
Representative grasps generated by our policy for each object are shown in Figure~\ref{fig:example-grasps-real}.

\section{Related Work}\label{sec:related-work}

Robotic grasping is normally approached either through analytical, model-based methods or data driven methods using either supervised or reinforcement learning.
The former focuses on constructing grasps that satisfy specific conditions, {\em e.g.},
gripper configuration, object contact points, force closure, task completion, {\em etc.}
while modelling the robot universe based on 3D models, partial meshes, and dynamic kinematic models~\cite{sahbani2012overview}.
Whereas the latter, learning-based methods, might learn from annotated datasets, or from the robot interacting with its
environment~\cite{lu2017planning,zeng2018learning}.
These learned grasping behaviors
tend to generalize better to unseen objects and situations.

Reinforcement Learning (RL) has been gaining prominence for robotic manipulation in recent years; many of these works have focused on learning grasping, but the majority focus on the simpler 2D gripper problem~\cite{johannink2018residual,thomas2018learning,yu2019sim,zeng2018learning,levine2016learning,quillen2018deep,caldera2018review,Fang2018Task}.
Andrychowicz {\em et al.} have trained a multi-finger robotic hand policy to repose a cube in-hand to match a desired pose~\cite{andrychowicz2018learning}.
Similar to our work they leverage simulation to train a policy to be deployed in the real world, however they do not focus on grasping, instead assuming the object already rests in the robot's hand.

The closest previous work to ours by Osa {\em et al.} also learns grasping policy for different grasping styles
using reinforcement learning~\cite{osa-ar2018-grasping} initialized by human demonstrations.
The grasping style is function of the surface mesh similarity to those seen during training and,
as such, wont be able to enforce a specific style {\em a priori}.

Another work with similar goals to ours uses supervised learning, coupled with analytical planning,
to plan multi-fingered grasps of different styles, {\em i.e.}, precision and power~\cite{lu2019modeling}.
They achieve this by explicitly modeling the grasp style as a decision variable in the grasp optimization.
Similar to previous robotics work~\cite{varley2015generating,mahler2016dex,lu2017planning,liu2019generating,kappler2015leveraging},
they learn a grasp success predictor from data.
Given a grasp configuration they use the gradients
from the predictor to refine the proposed grasp until it is predicted to be successful (has high probability).
Once the grasp configuration is found, it gets executed by a planner.
Our work differs from this as we seek to learn separate grasping policies for each grasp style
from a single human hand demonstration without relying on any planning algorithms  for grasp execution.
Other supervised-learning works have focus on grasping objects using one shot learning to predict contact points~\cite{kopicki2016one}.

Representation plays a very important role for learning in robotics manipulation.
Choosing the right one will allow for completion of learning downstream tasks.
Lee {\em et al.} proposed a method that learns an initial representation using unsupervised
learning methods \cite{lee2018making}.
Once the representation is learned they leverage the multi-sensing description to learn tasks
using RL methods, such as, peg-in-hole insertion.
Other work have explored using touching sensing to grasp objects under different assumptions, 
although very little work has been done on learning from using this sensor 
\cite{hsiao2010contact,laaksonen2012probabilistic,dang2014semantic,nikandrova2015category,calandra2018more}. 
Manuelli {\em et al.} also leverage keypoint representation to learn an agnostic representation
for a class of objects where a classical controller is written to accomplish a pick-and-place task~\cite{manuelli2019keypoints}.
Other works have focus on learning the full 6D pose of known objects for robotics pick and place~\cite{tremblay2018corl:dope,wang2019densefusion}.
Similar to \cite{platt2010null,chen2015adaptive}, our state representation also includes finger contact information to overcome shape and pose uncertainty; however,
they rely on hand-tuned, model-based controllers for execution. We believe our approach to be the first to explore using visual keypoints coupled with tactile-feedback in order to learn grasping behaviors with RL.

\section{Conclusion and Discussion}
\label{sec:conclusion}
We have presented a contextual policy search approach to learning policies for grasping unknown objects with multi-fingered hands using 
bounding box representation and contact sensing. 
We validate that our approach can train purely in simulation and be successfully deployed in the real world on a physical robot.
We introduce the use of bounding box keypoints as a contextual representation for the reward and, in turn, the policy.
We show that coupling this keypoint representation with contact sensing in the policy allows the robot to adapt to previously unseen shapes and overcome uncertainty in object pose estimation arising from noisy visual sensing.
This allows our method to handle objects with shape deviating greatly from that of a bounding box ({\em e.g.} a cone) we can optimize over the context variables to enable greater grasping performance without needing to retrain our learned policy. 
\section*{Acknowledgments}

The authors would like to thank Karl Van Wyk for his amazing help for setting up the 
the robotics system. We would also like to thank Nathan Ratliff, Rowland O'Flaherty, Ankur Handa, and 
Clemens Eppner for their technical help with various challenges. 



\bibliographystyle{IEEEtran}
\bibliography{example}

\end{document}